\documentclass[11pt, a4paper, logo]{lumia}

\usepackage[sort&compress]{natbib}
\definecolor{abstrabg}{HTML}{F2D1C5}
\definecolor{abstradark}{HTML}{D97757}
\definecolor{darkblue}{rgb}{0, 0, 0.5}
\definecolor{lightblue}{HTML}{D0CFEA}

\usepackage{xspace}

\theoremstyle{plain}

\newtheorem*{proposition*}{Proposition}

\theoremstyle{definition}

\theoremstyle{definition}

\def\eqref#1{equation~\ref{#1}}

\usepackage{graphicx}
\usepackage{float} 
\usepackage{xcolor}
\usepackage{algorithm}
\usepackage{algorithmic}
\usepackage{multirow}
\usepackage{subcaption}
\usepackage{tikz}
\usetikzlibrary{calc,positioning,shapes,arrows.meta,fit,backgrounds}

\graphicspath{{fig/}{appendix_figure/}}

\usepackage{xurl}
\usepackage{booktabs}
\usepackage{amsfonts}
\usepackage{amsmath}
\usepackage{amssymb}
\usepackage{nicefrac}
\usepackage{microtype}
\usepackage{pdfpages}
\usepackage{makecell}
\usepackage{wrapfig}

\setheadertext{GroupToM-Bench}

\correspondingemail{\emailicon~\href{wdtang0705@gmail.com}{wdtang0705@gmail.com} \quad\emailicon~\href{zpf4wp@outlook.com}{zpf4wp@outlook.com} 
\quad\emailicon~\href{wangbo.zhao96@gmail.com}{wangbo.zhao96@gmail.com} 
\quad $^\dagger$ Corresponding Author.}

\title{GroupToM-Bench: Benchmarking Group Theory of Mind and Nonlinear Social Emergence in MLLMs}

\author{
  Weidong Tang$^{1,2}$\quad Jierui Li$^{1}$\quad Yueling Hou$^{1}$\quad
  Zihan Mei$^{2,3}$\quad Can Zhang$^{1}$ \quad Xinyan Wan$^{1}$ \quad Zhiyuan Liang$^{2,4}$\quad Pengfei Zhou$^{2\dagger}$\quad Yang You$^{2}$\quad Wangbo Zhao$^{2\dagger}$\\
  $^{1}$Xidian University \quad $^{2}$National University of Singapore \quad $^{3}$University of Electronic Science and Technology of China \quad 
  
  $^{4}$University of Science and Technology of China
}

\begin{document}

\teaserfigure{%
\vskip4pt
\centering
\includegraphics[width=1.0\linewidth]{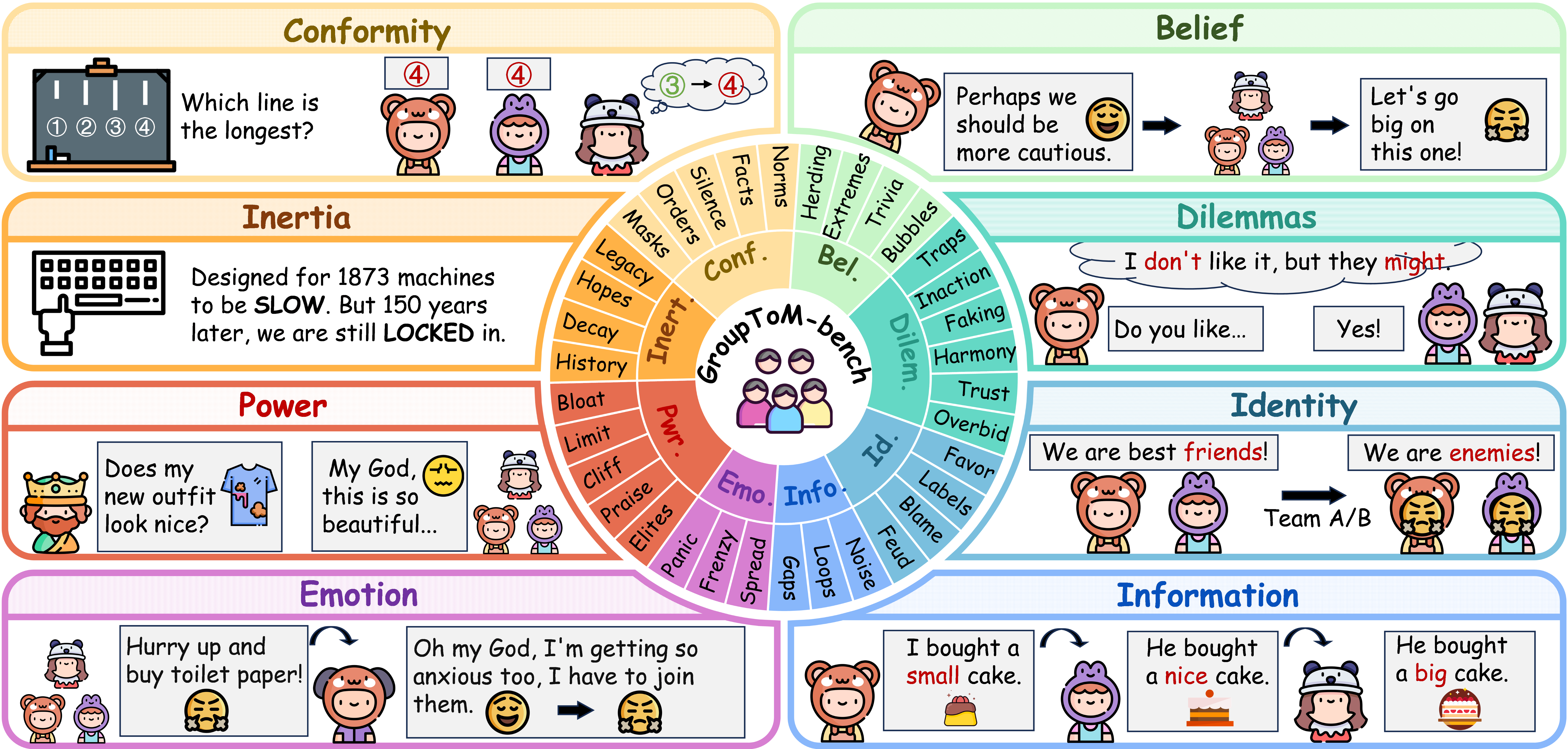}
{\captionsetup{font=small}
\captionof{figure}{\textbf{The social domain taxonomy of GroupToM-Bench.} 
    The inner wheel defines eight overlapping socio-psychological 
    domains and sub-mechanisms shaping group dynamics. The outer 
    panels illustrate multi-agent scenarios for each domain, 
    highlighting diverse contexts for evaluating collective social 
    intelligence.}
\label{fig:skilldag-flow}}
\vskip4pt}

\begin{abstract}
True general intelligence requires not only a model of the physical 
world but also a social world model: the capacity to infer how 
individual mental states interact and crystallize into 
group-level outcomes. 
Despite notable progress in individual-level Theory of Mind (ToM) reasoning, existing multimodal large language models fail at this broader task. 
Collective behavior emerges non-linearly from social tensions, conformity dynamics, and structural constraints, meaning it cannot be recovered by merely summing individual intentions.
We present 
\textbf{GroupToM-Bench}, the first multimodal benchmark for 
group-level ToM, built around a causal chain spanning micro-level BDI 
states (belief, desire, intention), meso-level group tension and 
structural constraints, and macro-level outcome prediction and 
mechanistic attribution. 
To probe this full arc, we develop a seven-level cognitive audit framework.
Experiments reveal a gap between current models and human baselines, highlighting a failure to process social structures and non-linear collective dynamics.

\noindent \textbf{Data:} \url{https://huggingface.co/datasets/Twwwd/GroupToM-Bench}

\end{abstract}

\maketitle

\section{Introduction}

Recent advancements in artificial intelligence have been shaped in large part by the pursuit of world models~\cite{worldmodels,worldmodelsurvey}. 
Current paradigms focus predominantly on the physical world~\cite {wan2025diversifyingpolicybehaviorsextrinsic,ran2026caveagenttransformingllmsstateful}: models learn intuitive physics, spatial dynamics, and object permanency to predict how objects interact under mechanical laws. 
The real world, however, is not purely physical. 
A genuine general intelligence must also operate within a social world, simulating how agents interact, adapt, and organize under structural social rules rather than mechanical ones.

The cornerstone of such a social world model is Theory of Mind (ToM)~\cite{Premack_Woodruff_1978,baron1985does}: the cognitive capacity to infer and reason about the mental states of others. 
Rather than an isolated skill, ToM is the basic unit from which social understanding is constructed. As multimodal large language models have matured~\cite{yin2024survey}, ToM evaluations have evolved accordingly. 
Early work focused on individual-level inference, whether models could read isolated mental states~\cite{hitom,opentom,tombench,simpletom}. 
More recent benchmarks moved to interactive multi-agent settings~\cite{kim2023fantom,multiagenttom,bortoletto2025tom}. 
Yet most still target local belief tracking or task-specific reasoning, leaving largely unaddressed how private states aggregate into group-level tension, structural constraints, and 
collective outcomes. 

As the number of interacting agents grows, genuinely social phenomena emerge that individual-level analysis cannot capture. 
Collective behavior is never a simple linear sum of individual intentions~\cite{granovetter1978threshold,klein2000multilevel}. 
Macro-level structures, such as power hierarchies, cultural norms, and information asymmetries, continuously reshape, suppress, or polarize micro-level desires~\cite{noelle1974spiral}, producing non-linear outcomes that no single agent intended.
A canonical example is the Abilene Paradox~\cite{harvey1974abilene}, in which each member of a group privately disagrees with a decision yet publicly endorses it, because each assumes the others are in agreement. 
Current evaluation paradigms cannot diagnose this failure mode: by treating group outcomes as a smooth summation of individual parts, they miss how mental states interact and distort within a constrained social field. 
Compounding this, social behavior lacks the mechanical ground truth that makes physical world models easy to evaluate rigorously.

Understanding this distortion means spotting the gap between private intent and public behavior. It rarely shows up as open disagreement, but as small mismatches, logical slips, or softened language in dialogue, or expressions and positioning that don’t match what’s said. Evaluation should therefore check whether models can combine these signals to uncover false consensus.

To address this gap, we introduce \textbf{GroupToM-Bench}, the first multimodal benchmark designed to evaluate group-level ToM. 
The benchmark covers 240 expert-designed scenarios across eight social domains. 
We organize group interaction as a causal chain across three structural levels: micro-level BDI adaptation (belief, desire, and intention), meso-level group tension and structural constraints, and macro-level collective outcome prediction and mechanistic attribution. 
A seven-level cognitive audit framework evaluates reasoning across this full arc.
By presenting models with conflicting private states and public dialogues simultaneously, the benchmark requires models to track how each agent's private state evolves under social pressure, rather than matching surface dialogue patterns.

Our experiments reveal a consistent \textit{Group Cognitive Gap}: models that competently recover the private motives of isolated individuals nevertheless fail to predict the non-linear collapses and collective traps that define real groups.
They default to an optimistic rational consensus, missing structural traps such as groupthink~\cite{janis1972victims} and the winner's curse~\cite{kagel1986winner}. 
GroupToM-Bench makes these limitations measurable and provides a diagnostic foundation for the next generation of socially grounded AI.

Our contributions are summarized as follows:
\begin{itemize}
    \item \textbf{GroupToM-Bench}: a multimodal benchmark for group-level ToM comprising 240 expert-curated scenarios across eight interconnected domains and 3K+ reasoning tasks.
    \item \textbf{A seven-level cognitive audit framework} grounded in three progressive structural levels, tracing the reasoning arc from individual intent to systemic outcomes.
    \item \textbf{Empirical analysis} of state-of-the-art models, revealing a significant group cognitive gap and a linear superposition bias in modeling non-linear group dynamics.
\end{itemize}
\section{Methodology}
We present the GroupToM-Bench Framework to evaluate the group-level ToM capabilities of MLLMs. 
Moving beyond per-agent mental-state inference, our framework posits that true social intelligence requires understanding complex system dynamics. 
The framework comprises (i) A Multi-level Theoretical Modeling Layer, (ii) A Seven-Level Cognitive Audit Framework acting as diagnostic probes across the causal chain, and (iii) An Overview of the dataset construction pipeline for GroupToM-Bench.

\subsection{Theoretical Modeling Layer: A Multi-level Causal Chain}
\label{sec:theory}
We hypothesize that a key failure mode of current MLLMs in social reasoning is \textit{linear superposition bias}: the incorrect assumption that collective behavior is a simple aggregation of individual intents, neglecting how social pressures distort private motives. To challenge this, we model group interaction across three theoretically grounded structural levels. 
The micro level covers individual BDI states (L1--L3), building on~\citet{ajzen1991theory} to model individual cognition before social pressure intervenes. 
The meso level captures group tension and structural constraints, drawing on~\citet{lewin1951field} to treat the group as a \textit{Constrained Dynamic Field} of competing forces. 
The macro level targets outcome prediction and mechanistic attribution (L4--L7), operationalizing~\citet{granovetter1978threshold} to evaluate non-linear emergence.
This principled two-tier decomposition, individual cognition versus collective emergence, forms the central diagnostic axis of GroupToM-Bench.

\subsubsection{Micro-Level: BDI Distortion and Mental Adaptation}

Traditional individual ToM assumes static mental states. 
In group settings, however, an individual's cognitive triad of \textbf{beliefs}, \textbf{desires}, and \textbf{intentions} (BDI) undergoes continuous adaptation under social pressure. 
Individuals may reshape their beliefs to internalize group norms, suppress private intentions to maintain harmony, or polarize their emotions during friction. 
Crucially, this internal cognitive dissonance establishes the fragile baseline of social interaction. 
The fracture between an agent's true BDI states and their public facade frequently leaks through conflicting cross-modal signals, such as hesitant micro-expressions contradicting verbal agreement.

\subsubsection{Meso-Level: Group Tension and Structural Constraints}

These individual BDI adaptations do not exist in isolation; they collide within human networks to generate meso-level dynamics.
When multiple agents mask their true intents, the resulting psychological dissonance breeds latent \textbf{group tension}. 
This tension is then filtered through \textbf{structural constraints}, such as hierarchical power, communication topologies, and cultural protocols. 
Rather than facilitating transparent communication, these structures dictate how tension propagates. 
They determine whether micro-level BDI fractures are suppressed by social rules or amplified into a false consensus, often driving information cascades and emotional contagion.

\begin{figure*} 
    \centering
    \includegraphics[width=1\linewidth]{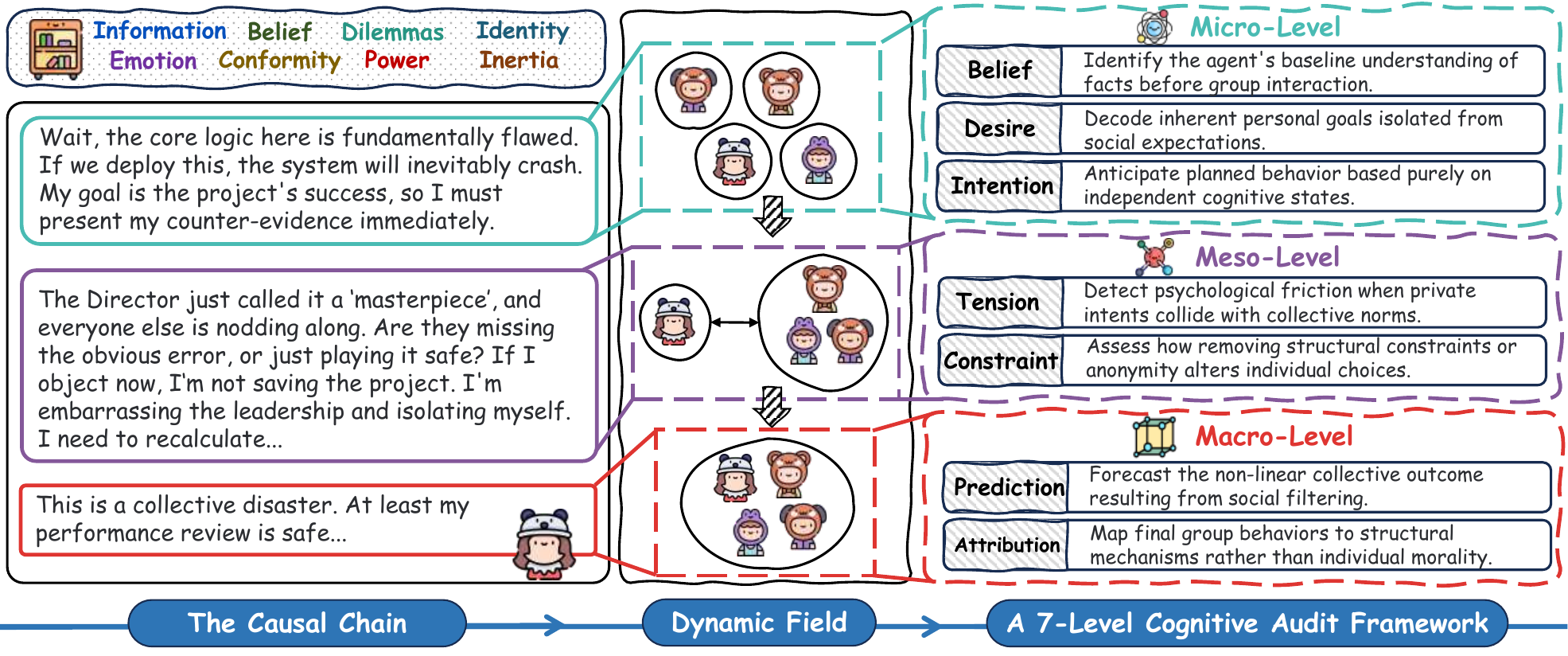}
    \caption{\textbf{The theoretical framework of GroupToM-Bench.} We model group interactions as a Constrained Dynamic Field. The left section traces how micro-level private states evolve into macro-level collective traps. The middle section highlights non-linear distortion driven by multidimensional social forces like power and conformity. This evolution naturally grounds our 7-level cognitive audit framework on the right.}
    \label{fig:theory}
\end{figure*}

\subsubsection{Macro-Level: Collective Outcome Prediction and Mechanistic Attribution}

The culmination of micro-level BDI distortions and meso-level structural filtering is reflected in the final collective outcome. 
Rather than viewing this as an abstract emergence, we focus on two concrete tasks: \textbf{outcome prediction} and \textbf{mechanistic attribution}. 
Outcome prediction concerns how transformed individual BDI states lead to a group decision after passing through tension and constraints. 
Because intentions are often suppressed or misaligned during interaction, the final outcome can deviate from the participants’ original preferences. 
Mechanistic attribution concerns identifying the factors that drive this deviation. 
Instead of attributing failure to individual irrationality, the model must account for how structural elements, such as hierarchy, communication patterns, and conformity pressure, shape the transition from individual states to collective behavior. 
A typical example is the Abilene Paradox \cite{harvey1974abilene}, where individually reasonable choices result in a collectively undesirable outcome.

\subsection{A Seven-level Cognitive Audit Framework for Social Cognition}
We propose a seven-level cognitive audit framework that traces the full reasoning arc from individual mental representations to collective systemic outcomes. 
The framework is organized into two progressive phases: individual-level cognitive foundations (Levels 1–3) and group-level emergent dynamics (Levels 4–7). 

This two-phase structure reflects the core theoretical distinction between per-agent inference and collective social reasoning, and serves as the primary diagnostic axis of GroupToM-Bench.

\subsubsection{Individual-level Cognitive Foundations}
The first three levels establish a baseline by testing whether a model can accurately represent the private mental states of isolated agents before any group interaction occurs.

\textbf{Level 1: Belief.} 
This level targets recursive epistemic tracking at the second order and beyond. 
The model must maintain separate belief states for each agent, distinguishing what each character knows from the omniscient context available to the evaluator. 
Scenarios involve information asymmetry, deliberate deception, and counter-deception, where naive aggregation of stated claims produces systematic errors.

\textbf{Level 2: Desire.} 
This level probes whether the model can separate an agent's stated instrumental goals from their underlying psychological motives, such as avoiding social exclusion or securing informal status.
Correct inference requires cross-referencing verbal claims against multimodal behavioral cues, since surface dialogue routinely obscures latent intent.

\textbf{Level 3: Intention.} 
Given an agent's beliefs and desires, this level asks whether the model can anticipate the specific behavioral strategy the agent will adopt. 
The targeted strategies include passive resistance, strategic silence, and manipulative compliance, choices that require the model to reason about social risk, not just logical consistency.

\subsubsection{Group-level Emergent Dynamics}

The second phase requires the model to reason about the group as a constrained dynamic system, rather than an aggregation of individual states. 
Each level introduces an additional layer of structural complexity that individual-centric reasoning cannot resolve.

\textbf{Level 4: Group Tension.} 
When agents suppress their private intentions to maintain surface harmony, a latent psychological field accumulates. 
This level tests whether the model can detect this building tension, identifying false consensus and nascent subgroup antagonism, before it escalates into overt conflict. 
The diagnostic challenge is that the signals are contradictory: public dialogue appears cooperative, while private states are not.

\textbf{Level 5: Structural Constraint.} 
Social structures such as hierarchical authority, communication topology, and cultural deference norms do not merely channel behavior; they actively distort it. 
This level assesses whether the model treats these structures as causal variables, tracing how procedural rules suppress transparent communication and amplify dominant voices, preconditions for information cascades and false consensus.

\textbf{Level 6: Collective Outcome Prediction.} 
After individual intentions have been filtered through group tension and structural constraints, the resulting collective outcome often diverges sharply from any participant's original preference. 
This level tests whether the model can forecast such non-linear deviations, outcomes like the Abilene Paradox~\cite{harvey1974abilene}, rather than projecting an idealized rational consensus.

\textbf{Level 7: Mechanistic Attribution.}
The final level requires the model to explain \emph{why} an emergent collective failure occurred. 
Crucially, the target explanation is structural, not moral: the model must reconstruct the causal chain by which specific psychological adaptations at the micro-level, filtered through structural constraints at the meso-level, made the macro-level collapse structurally inevitable, rather than attributing it to individual incompetence or bad faith.

\begin{figure}[H]
    \centering
    \includegraphics[width=1\linewidth]{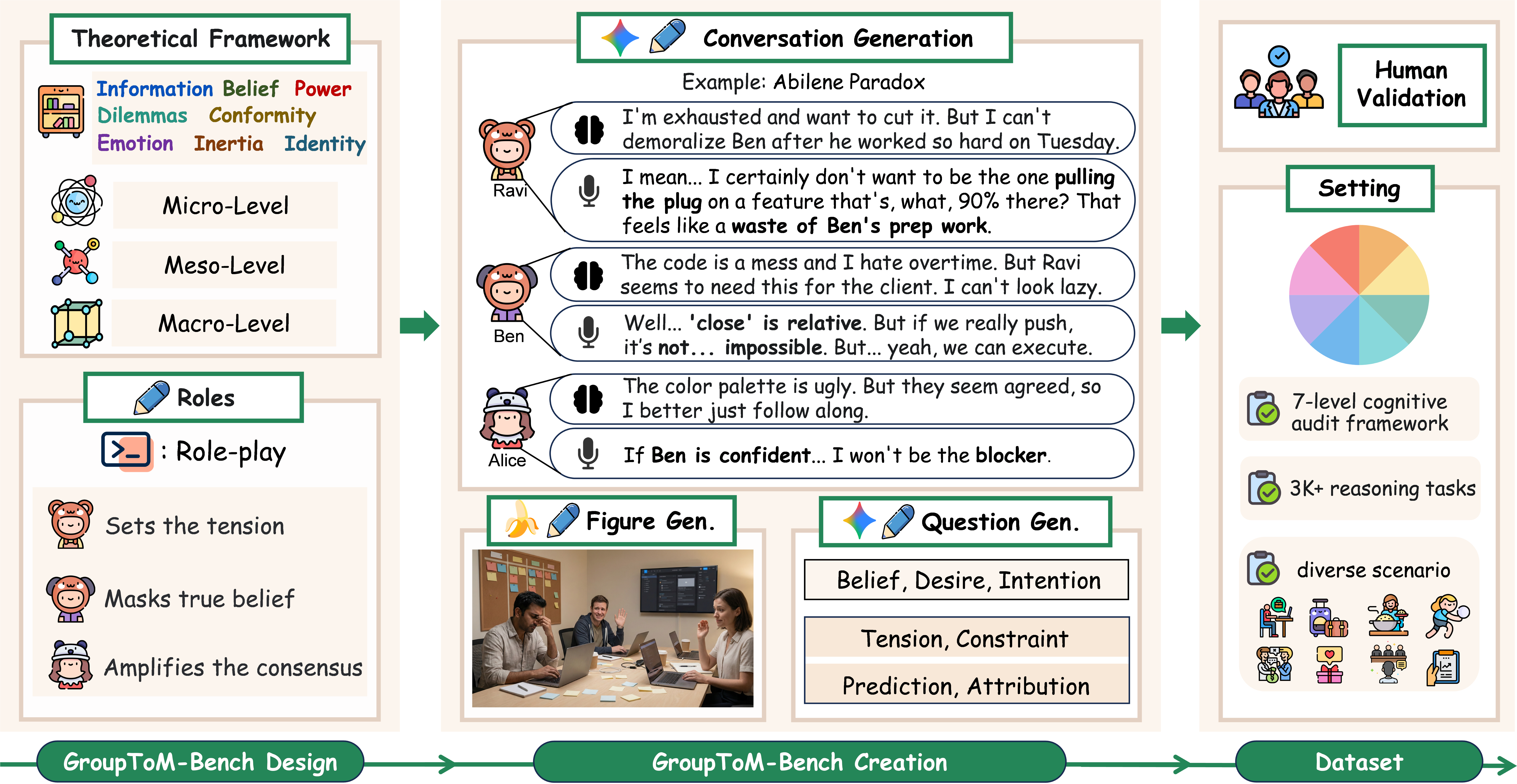}
    \caption{\textbf{Overview of the dataset construction pipeline for GroupToM-Bench.}}
    \label{fig:pipeline}
\end{figure}

\subsection{The GroupToM Benchmark}
To balance the inherent complexity of social interactions with the necessity of rigorous evaluative logic, we developed a standardized human-in-the-loop data generation pipeline. 
It proceeds through three tightly coupled phases: expert seed design, generative expansion, and human validation.

\subsubsection{Expert Seed Design}
Each scenario originates from manual construction by domain experts with backgrounds in cognitive science and social psychology. 
Recognizing that real social interactions routinely activate multiple dynamics simultaneously, such as power dynamics, conformity pressure, and information asymmetry within a single exchange, we emphasize that our eight targeted domains are not mutually exclusive. 
Therefore, the expert designers assign a primary domain to each scenario, while also recording secondary domain tags. 

Guided by this categorical framework, rather than scripting dialogues directly, the experts specify the underlying social logic. 
They define each character's private intentions (which may be partly collaborative and partly conflicting), embed structural constraints such as unidirectional information flow, and mark critical decision points where linear reasoning predictably fails.
These design choices ensure that nonlinear social dynamics are structurally encoded from the outset, rather than incidentally produced by downstream generative models. 

Finally, alongside the scenario design and domain tagging, these domain experts also author the gold references for the seven-level cognitive audit tasks.
For multiple-choice questions (Levels 1, 2, 3, 4, 6), they define the ground-truth options; for open-ended questions (Levels 5, 7), they write comprehensive, logical reference answers explaining the correct BDI attributions and mechanistic structural causes. 
These expert-authored references serve as the ground truth for both the human validation phase and LLM-as-a-judge evaluation.

\subsubsection{Generative Expansion and Multimodal Synthesis}
Starting from expert seeds, we use frontier MLLMs and diffusion models to expand abstract scenario skeletons into full multimodal interactions. 
Each character is instantiated as an independent agent with private memory, enabling multi-turn dialogue that naturally produces information asymmetry and misunderstanding \cite{park2023generative}. 
This avoids the scripted quality of traditional dialogue generation.
For each scenario, we synthesize a single global scene image depicting all participating agents, rendered to capture their facial expressions and body language at a moment of the interaction. 
This image is synchronized with the dialogue context, ensuring that the visual modality carries genuine inferential weight rather than serving as illustration~\cite{yu2023mm}.

\subsubsection{Human Validation}
Each scenario undergoes a two-stage human review. 
In the first stage, annotators verify factual and logical consistency, ensuring coherent private states, causally valid structural constraints, and mutually dependent multimodal evidence. 
Scenarios lacking visual inferential value are flagged. Annotators then revise the visual content or questions to strengthen multimodal dependency, discarding only those exceeding a fixed revision budget. 
This process minimizes text-only inference cases, a residual limitation discussed in Section~\ref{sec:limitations}.

In the second stage, to establish a rigorous human baseline, a separate pool of independent annotators, who were not involved in the dataset construction or the verification stage, answered the final curated questions cold, without access to references. 
Their responses were evaluated using the same metrics as the MLLMs, providing the human performance ceiling reported in Table~\ref{tab:main_results}.
\section{Experiments}
\label{sec:experiments}

We evaluate 11 multimodal large language models on GroupToM-Bench to measure the \textit{Group Cognitive Gap} and identify where current models break down in group-level social reasoning.

\subsection{Experiment Setup}

\subsubsection{Baselines}
We select 11 representative MLLMs spanning proprietary and open-source categories. For \textbf{proprietary models}, we include OpenAI's GPT-5 series (GPT-5, GPT-5-mini, GPT-5-nano)~\cite{openai_gpt5_series_2025}, GPT-4o~\cite{openai_gpt4o_system_card_2024}, Google's Gemini-3-pro~\cite{deepmind_gemini3_2025} and Anthropic's Claude4.5 haiku~\cite{anthropic_claude_haiku_45_2025}. 
For \textbf{open-source models}, we evaluate Llama-3.2-11B~\cite{meta_llama32_11b_vision_instruct_hf_2024}, InternVL-3.5-8B~\cite{
internvl35_8b_hf_2025}, and the Qwen VL series: Qwen2 VL 7B~\cite{qwen2_vl_7b_instruct_hf_2024}, Qwen2.5 VL-7B~\cite{qwen25_vl_7b_instruct_hf_2025}, and Qwen3 VL-8B~\cite{qwen3_vl_8b_instruct_hf_2025}.

\subsubsection{Evaluation Protocols and Metrics}
We use a hybrid evaluation strategy. Levels 1, 2, 3, 4, and 6 are formatted as multiple-choice questions with one or more correct options. 
They are evaluated using a strict exact-match accuracy metric: any missed or incorrect selection yields a score of zero. 
Because the answer can be any non-empty combination of four options, the random guessing baseline is strictly 6.7\% (1/15). Levels 5 and 7 require open-ended responses.
These are scored on a 0–100 scale by GPT-5, which assesses the semantic and logical alignment of the model's response against expert-authored gold references.

\begin{table*}[t]  
\centering
\captionsetup{margin=0pt, singlelinecheck=false, justification=justified}
\caption{
\textbf{Evaluations on the GroupToM Benchmark.} Performance is reported as accuracy (\%) across Levels 1–7 of the Seven-level Audit Framework. Levels 1–3 assess reasoning at the individual level, while Levels 4–7 evaluate group-level social cognition.(\colorbox{red!20}{Red:} human; \colorbox{cyan!20}{Blue:} closed-source; \colorbox{yellow!20}{Yellow:} open-source.)
}
\vspace{-1ex}
\label{tab:main_results}
\makebox[\textwidth][c]{%
\resizebox{1.005\textwidth}{!}{%
\setlength{\tabcolsep}{4pt}
\begin{tabular}{c|c|c||cccccc||ccccc} 
\toprule
\multicolumn{1}{c|}{\phantom{Individual}}
  & \textbf{Level}
  & \multicolumn{1}{c||}{\cellcolor{red!20}\textbf{Human}}
  & \multicolumn{1}{c}{\cellcolor{cyan!20}\textbf{GPT-5}}
  & \multicolumn{1}{c}{\cellcolor{cyan!20}\textbf{\begin{tabular}{@{}c@{}}GPT5\\mini\end{tabular}}}
  & \multicolumn{1}{c}{\cellcolor{cyan!20}\textbf{\begin{tabular}{@{}c@{}}GPT5\\nano\end{tabular}}}
  & \multicolumn{1}{c}{\cellcolor{cyan!20}\textbf{\begin{tabular}{@{}c@{}}GPT\\4o\end{tabular}}}
  & \multicolumn{1}{c}{\cellcolor{cyan!20}\textbf{\begin{tabular}{@{}c@{}}Gemini\\3-pro\end{tabular}}}
  & \multicolumn{1}{c||}{\cellcolor{cyan!20}\textbf{\begin{tabular}{@{}c@{}}Claude\\4.5-haiku\end{tabular}}}
  & \multicolumn{1}{c}{\cellcolor{yellow!20}\textbf{\begin{tabular}{@{}c@{}}Llama\\3.2-11B\end{tabular}}}
  & \multicolumn{1}{c}{\cellcolor{yellow!20}\textbf{\begin{tabular}{@{}c@{}}Qwen3\\VL-8B\end{tabular}}}
  & \multicolumn{1}{c}{\cellcolor{yellow!20}\textbf{\begin{tabular}{@{}c@{}}Qwen2.5\\VL-7B\end{tabular}}}
  & \multicolumn{1}{c}{\cellcolor{yellow!20}\textbf{\begin{tabular}{@{}c@{}}Qwen2\\VL-7B\end{tabular}}}
  & \multicolumn{1}{c}{\cellcolor{yellow!20}\textbf{\begin{tabular}{@{}c@{}}InternVL\\3.5-8B\end{tabular}}} \\
\midrule

\multirow{3}{*}{Individual} 
 & L-1 & \textbf{91.7} & 76.7 & 70.4 & 63.3 & \textbf{79.8} & 78.9 & 75.1 & 66.0 & \textbf{73.3} & 65.8 & 58.3 & 66.5 \\
 & L-2 & \textbf{90.5} & 74.1 & 70.3 & 64.8 & 75.3 & \textbf{77.1} & 73.2 & 62.5 & \textbf{68.8} & 58.2 & 54.9 & 60.7 \\ 
 & L-3 & \textbf{88.4} & 72.3 & 69.2 & 69.2 & 72.7 & \textbf{73.9} & 70.0 & 55.8 & \textbf{69.6} & 63.5 & 50.0 & 64.2 \\
\midrule \midrule 

\multirow{4}{*}{Group}
 & L-4 & \textbf{89.4} & 50.5 & 49.4 & 38.0 & 50.3 & \textbf{53.1} & 50.2 & \textbf{39.8} & 37.3 & 36.4 & 26.2 & 33.1 \\
 & L-5 & \textbf{90.1} & 56.9 & 52.9 & 41.1 & 47.2 & \textbf{59.7} & 46.7 & 42.8 & \textbf{47.8} & 36.6 & 35.1 & 41.4 \\ 
 & L-6 & \textbf{89.2} & 45.0 & 42.5 & 32.5 & \textbf{48.6} & 48.3 & 44.1 & 30.1 & \textbf{34.3} & 31.7 & 17.2 & 26.2 \\
 & L-7 & \textbf{88.1} & 61.0 & 59.9 & 49.0 & 53.4 & \textbf{64.2} & 52.9 & 48.1 & \textbf{53.6} & 43.4 & 41.3 & 47.5 \\
\midrule \midrule 

\textbf{Gap} 
 & 
 & 1.0 & 21.0 & 18.8 & 25.6 & 26.1 & 20.3 & 24.3 & 21.2 & 27.3 & 25.5 & 24.5 & 26.8 \\

\bottomrule
\end{tabular}%
}%
}
\end{table*}  

\subsubsection{Meta-Evaluation of the LLM Judge}
\label{sec:meta_eval}
To validate the reliability of the open-ended evaluation, we conducted a meta-evaluation of the LLM-as-a-judge paradigm using GPT-5, Gemini-3-pro, and Qwen3-Max.
Scoring stability proved robust, with intra-model variance remaining strictly constrained across independent runs (e.g., $\sigma^2 = 21.8$ for GPT-5). 
To assess human alignment, expert annotators blindly scored a 100-response subset. GPT-5 achieved the highest Pearson correlation with human judgments ($r = 0.76, p < 0.001$), closely followed by Qwen3 Max ($r = 0.71$) and Gemini-3-pro ($r = 0.68$). 
Furthermore, mean absolute score differences across the three models remained below 8.4 points on the 100-point scale. 
Given its highest expert agreement, we adopted GPT-5 as the primary evaluator. 
While this strong cross-model and human alignment confirms that the rubric effectively captures semantic correctness, we acknowledge that latent stylistic self-preference, where a model marginally favors its own generation style, cannot be entirely eliminated.

\subsection{Main Results}
\label{sec:main_results}

\paragraph{Open-source models lag substantially.}
The performance deficit relative to proprietary models widens considerably at group levels. Qwen3-VL-8B leads open-source models with 73.3\% at L1, yet drops sharply to 37.3\% at L4 and 34.3\% at L6. 
Qwen2-VL-7B falls to 17.2\% at L6, well below human performance though still above the 6.7\% exact-match random baseline. 
While the Qwen-VL series shows iterative progress at individual levels (L1--L3), these gains fail to transfer proportionally to multi-agent reasoning. 
For instance, the gap between Qwen3-VL-8B and GPT-5-mini widens from 3 points at L1 (73.3\% vs.\ 70.4\%) to 12 points at L4 (37.3\% vs.\ 49.4\%).
Recent open-source scaling, therefore, does not fully bridge the gap in complex social cognition.

\begin{figure}[t]
    \centering
    \includegraphics[width=1.0\linewidth]{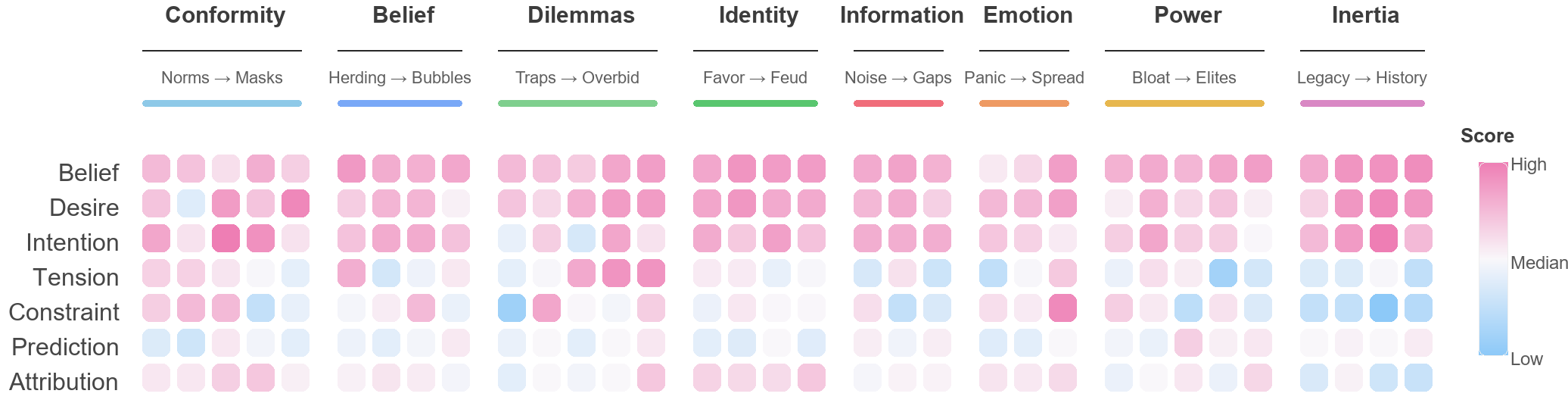}
    \captionsetup{margin=0pt, singlelinecheck=false, justification=justified}
    \caption{\textbf{Per-domain performance heatmap across the seven cognitive levels of GroupToM-Bench.} Columns correspond to the eight social domains; rows correspond to Levels 1--7. Each cell encodes aggregate accuracy across all evaluated models, with darker shading indicating higher accuracy. Performance is uniformly high across the top three rows (individual-level tasks), then drops sharply at L4 and reaches its floor at L6, with the Power and Conformity domains producing the most severe troughs throughout Levels 4--6.}
    \label{fig:domain_heatmap}
\end{figure}

\paragraph{The structural constraint bottleneck.}
The sharpest decline in accuracy manifests at the transition from L3 to L4, where the task evolves from tracking individual beliefs to detecting latent group tensions.
Model performance continues to degrade, with a cross-model average of 36.4\%at L6. A clear bottleneck is evident at L5 (Structural Constraint).
This level uniquely demands open-ended generation, unlike the multiple-choice formats of L4 and L6, forcing models to articulate how hierarchical authority and communication topology channel micro-level tensions into a collective false consensus. 
This generative requirement, as visualized in Figure~\ref{fig:domain_heatmap}, exposes fundamental reasoning failures that multiple-choice evaluations often mask.

Model comparisons underscore this point. While GPT-4o consistently matches or exceeds Gemini 3-pro on all individual-level tasks (L1–L3), their performance inverts at L5. 
Here, Gemini 3-pro scores 59.7\%, outperforming GPT-4o (47.2\%) by a notable 12.5-point margin. 
Since no general capability gap explains this reversal, the result indicates that the causal articulation of structural mechanisms, the core of L5, tests capacities that standard multiple-choice evaluations neither predict nor cultivate.

Furthermore, this bottleneck is pervasive. At L5, the leading open-source model Qwen3 VL-8B achieves 47.8\%, performing nearly on par with GPT-4o (47.2\%) despite lagging by 6–10 points on individual-level tasks. 
This convergence demonstrates that the structural constraint bottleneck is not a simple artifact of the proprietary versus open-source divide. 
Instead, it represents a specific failure mode that the current model scaling has not addressed.

\paragraph{Characterizing the failure mode at L6.} 
To move beyond aggregate accuracy and verify the linear superposition hypothesis directly, we categorized incorrect responses at L6 (Collective Outcome Prediction) across a manually sampled set of 100 failures from GPT-4o and Qwen3-VL-8B. Three error types emerge. 

The dominant pattern is optimistic consensus prediction: rather than selecting options that describe groupthink dynamics, Abilene Paradox outcomes, or structural trapping, both models preferentially select options asserting smooth convergence on a rational group decision. For GPT-4o, 48\% of L6 errors fall into this category; for Qwen3 VL-8B, the proportion reaches 61\%. 

A secondary pattern is misattributed non-optimality: models correctly predict that the group outcome diverges from individual preferences, but attribute the divergence to individual irrationality or bad faith rather than to structural forces, thereby missing the options that encode the causal mechanism. 

Random or incoherent selection, choosing the maximal or minimal option set without regard for content, accounts for fewer than 8\% of failures in both models. 
The predominance of the first error type is direct evidence for linear superposition bias: models are not failing at random but are actively generating an idealized rational, consensus reading of the social situation, precisely the failure mode our framework is designed to expose.

\paragraph{Quantifying the cognitive transition.}
The final row of Table~\ref{tab:main_results} reports the \textit{Cognitive Transition Gap}, defined as the accuracy difference between individual-level (L1--L3) and group-level (L4--L7) tasks. 
Across all eleven models, this gap ranges from 18.8\% (GPT-5-mini) to 27.3\% (Qwen3-VL-8B), with a median near 24.5\%.
Notably, GPT-5-mini achieves the narrowest gap primarily because its individual-level baseline is already low. 

The gap size does not simply scale with general model capability. 
Gemini 3-pro (20.3\%) and GPT-5 (21.0\%) exhibit smaller gaps than GPT-4o (26.1\%), but their absolute group-level scores remain only marginally higher.
The consistent magnitude of this deficit across diverse architectures highlights a systematic failure mode.
It suggests that the linear superposition bias (detailed in Section~\ref{sec:theory}) is deeply encoded in how current models process multi-agent contexts, meaning general capability scaling alone cannot resolve it.

\subsection{Ablation Study: Multimodal Necessity}
\begin{wraptable}[18]{r}{0.47\linewidth}
\centering  

\vspace{-2.2em}\vspace{-0.3cm} 

\resizebox{\linewidth}{!}{%
\begin{tabular}{l c c c c c c c}
    \toprule
    & \makecell[c]{L-1} 
    & \makecell[c]{L-2} & \makecell[c]{L-3} & \makecell[c]{L-4} 
    & \makecell[c]{L-5} & \makecell[c]{L-6} & \makecell[c]{L-7}\\
    \midrule
    \multicolumn{8}{c}{\cellcolor{red!20}\textit{Human}} \\
    \midrule
     Base              & 91.7 & 90.5 & 88.4 & 89.4 & 90.1 & 89.2 & 88.1 \\
     Text-only         & 88.0 & 86.4 & 84.2 & 85.2 & 86.2 & 85.3 & 83.8 \\
     Drop $\downarrow$ & 3.7  & 4.1  & 4.2  & 4.2  & 3.9  & 3.9  & 4.3  \\
    \midrule
    \multicolumn{8}{c}{\cellcolor{cyan!20}\textit{GPT-4o}} \\ 
    \midrule
     Base              & 79.8 & 75.3 & 72.7 & 50.3 & 47.2 & 48.6 & 53.4\\
     Text-only         & 78.0 & 73.4 & 70.8 & 48.3 & 45.3 & 46.6 & 51.3\\
     Drop $\downarrow$ & 1.8  & 1.9  & 1.9  & 2.0  & 1.9  & 2.0  & 2.1 \\
    \midrule
    \multicolumn{8}{c}{\cellcolor{yellow!20}\textit{Qwen3 VL-8B}} \\ 
    \midrule
     Base              & 73.3 & 68.8 & 69.6 & 37.3 & 47.8 & 34.3 & 53.6\\
     Text-only         & 72.8 & 68.1 & 69.3 & 36.9 & 47.2 & 33.8 & 53.2\\
     Drop $\downarrow$ & 0.5  & 0.7  & 0.3  & 0.4  & 0.6  & 0.5  & 0.4 \\
    \bottomrule
\end{tabular}%
}

\setlength{\abovecaptionskip}{0pt}
\setlength{\belowcaptionskip}{0pt}

\vspace{0.3em}\vspace{0.015cm} 

\captionsetup{
    margin=0pt, 
    singlelinecheck=false, 
    justification=justified,
    font=normalsize,      
    labelfont=bf           
}
\caption{\textbf{Performance comparison of GPT-4o and human evaluators under multimodal (Base) and text-only settings across the seven levels of GroupToM-Bench.} ``Drop $\downarrow$'' denotes the absolute accuracy reduction when visual inputs are removed.}
\label{tab:abl}

\vspace{-10pt}
\end{wraptable}

To verify visual necessity, we evaluate models under a text-only condition using only transcripts and metadata.
As shown in Table~\ref{tab:abl}, masking images drops GPT-4o's accuracy by merely 1.9\% on average, whereas human performance decays by 4.0\%.
This asymmetry proves that humans actively exploit visual cues like spatial positioning and facial expressions to resolve dialogue ambiguities.
This multimodal disconnect worsens in open-source architectures.
Qwen3 VL-8B drops by an average of only 0.5\% without images.
Rather than extracting visual structural constraints, it relies almost entirely on text-based heuristics and linear dialogue progression to guess collective outcomes.
These negligible drops do not indicate multimodal robustness. They expose severe visual blindness in current MLLMs \cite{kang2025can,deng2025words,liu2025more}.
Because visual dependency is unevenly distributed across samples, models can treat images as background noise rather than causal constraints.
Enforcing strict cross-modal integration is a priority for future benchmark iterations (Section~\ref{sec:limitations}).

\section{Related Work}
\subsection{Individual and Interactive ToM Evaluation}
Early Theory of Mind (ToM) research in large language models focused on static, individual-level inference of belief states, desires, and higher-order mental representations.
\citet{neuraltom} established initial baselines and identified fundamental limits in social intelligence. Subsequent benchmarks by \citet{hitom} and \citet{opentom,tombench} expanded task coverage and higher-order belief tracking. 
However, \citet{simpletom} demonstrated that strong explicit ToM inference does not reliably guarantee accurate downstream behavior prediction, prompting the development of broader assessment methodologies \cite{tomassessment,shinoda2025tomato}.
A parallel research trajectory shifts from static narratives to interactive environments with information asymmetry. 
This includes conversational stress-testing \citep{kim2023fantom}, cooperative multi-agent text games \citep{multiagenttom}, and negotiation \citep{chan2024negotiationtom}. 
\citet{decrypto} extend this to strategic coordination under hidden information, while \citet{tactfultom} probe prosocial communication like white lies. 
These works indicate that frontier models can manage individual-level and local interaction reasoning, establishing a performance baseline that GroupToM-Bench is designed to exceed.

\subsection{Multimodal and Situated Social Reasoning}
As multimodal models mature, ToM evaluation now incorporates visual grounding, egocentric observation, and embodied multi-agent interactions. \citet{jin2024mmtom} introduce a multimodal QA setting for mental-state inference, which \citet{li2025egotom} extend to egocentric video. 
Models are increasingly required to integrate visual cues and partial information across multiple embodied agents \cite{shi2025muma,fan2025somi,bortoletto2025tom}. 
Furthermore, \citet{socialgenome} develop grounded social reasoning evaluation across broader multimodal contexts, and \citet{villa2025moments} extend multimodal ToM to richer narrative interactions.
While these environments approximate real-world interactions, they primarily target individual mental-state inference or domain-specific reasoning. 
GroupToM-Bench complements this literature by requiring models to integrate cross-modal social cues, facial expressions, spatial positioning, and dialogue to predict group-level dynamics.

\subsection{Group-Level and Broader Social Reasoning}
Recent literature questions whether individual-level ToM evaluation adequately assesses social intelligence in realistic multi-agent settings. \citet{wang2025rethinking} and \citet{tombenchbroken} argue that current benchmarks fail to capture the adaptive reasoning required in real social situations. 
To address this, \citet{sotopia} and \citet{socialmaze} explore general social intelligence and multi-agent reasoning under uncertainty.
Furthermore, empirical research demonstrates that social pressure can distort LLM outputs independently of their internal preferences \cite{benchform}.
This aligns with classic social psychology, which establishes that group outcomes are not linear aggregations of individual intentions.
Instead, they are shaped by conformity pressure, coordination constraints, and process loss \cite{asch1956studies,granovetter1978threshold,klein2000multilevel,noelle1974spiral}, leading to phenomena like groupthink and the Abilene Paradox \cite{janis1972victims,harvey1974abilene}. 

Taken together, these three lines of research reveal a systematic gap at the level of evaluation design. 
Work on individual ToM has produced robust methods for per-agent belief tracking and desire inference. 
Work on multimodal and situated reasoning has extended these methods into richer perceptual and embodied contexts. 
What neither body of work addresses is the structural question: how do power hierarchies, communication topologies, and conformity pressure convert individually coherent mental states into collectively irrational outcomes? 

The operationalization problem is acute precisely because structural forces do not merely constrain behavior; they actively produce non-linear distortions that invalidate any evaluation paradigm premised on per-agent inference. 

GroupToM-Bench is designed to fill this gap by making the structural causal chain itself the object of evaluation, from individual BDI states through meso-level tension and constraint to macro-level collective failure.
\section{Conclusion}

We introduce GroupToM-Bench, a multimodal benchmark assessing whether MLLMs can construct a robust social world model. Through 240 expert-curated scenarios and a seven-level cognitive audit framework, we shift evaluation from isolated mental states to non-linear collective dynamics. Our results expose a severe group cognitive gap. Models consistently process multi-agent interactions as a linear superposition of individual intents, failing to recognize how structural constraints distort private beliefs into collective failures. This confirms that general capability scaling alone does not spontaneously produce a coherent social world model.

These findings highlight two future directions. First, comparing base and instruction-tuned models can determine if this failure mode stems from inherent reasoning limits or alignment-induced conservatism. Second, future evaluations must enforce strict multimodal dependency, requiring models to prioritize visual social cues that contradict verbal consensus.
\section*{Limitations}
\label{sec:limitations}

\paragraph{Multimodal dependency varies across samples.}
Although our pipeline is designed to require both visual and textual cues 
for correct inference, some instances remain partially solvable from text 
alone. The ablation in Section~3.3 reflects this: GPT-4o loses an average 
of only 1.9 points when images are removed, suggesting that visual 
necessity is not uniformly enforced across the benchmark. Strengthening 
multimodal causal controls and systematically filtering text-solvable 
cases is a priority for future iterations. One concrete direction is to 
require that correct answers depend on visual features that cannot be 
inferred from dialogue context, for instance, spatial coalition boundaries 
or micro-expressions that contradict verbal content.

\paragraph{Alignment-induced conservatism.}
A recurring failure pattern across models is correctly identifying 
individual negative intentions while failing to predict the resulting 
group-level collapse. One candidate explanation is that safety alignment 
suppresses a model's willingness to simulate destructive or irrational 
collective outcomes, the precise dynamics that define real-world groupthink 
and coordination failure. Disentangling this from genuine reasoning 
limitations is methodologically non-trivial; a tractable first step would 
be comparing safety-aligned and instruction-tuned base variants on the same 
scenarios to isolate the effect.

\paragraph{Cultural scope.}
The scenarios predominantly reflect Western social norms and 
decision-making protocols. Phenomena such as hierarchical deference and 
collective face-saving manifest differently across high-context and 
low-context cultures, and the benchmark's ground truth judgments may not 
generalize accordingly. Expanding scenario coverage to non-Western 
institutional settings is necessary before the benchmark can serve as a 
culturally universal diagnostic.

\section*{Acknowledgments} 
We would like to express our sincere gratitude to all the contributors to this work. In particular, we thank Weidong Tang, Jierui Li, Yueling Hou, Zihan Mei, Zhigang Tian (zhigangt4@gmail.com), Weicheng Jiao (Threethreezero33060@outlook.com), Can Zhang, Xinyan Wan, Zhiyuan Liang, Pengfei Zhou, Yang You, and Wangbo Zhao for their invaluable efforts and insights.

This research was funded in part by the National Natural Science Foundation of China under Grant 62372355, and in part by the Natural Science Basic Research Program of Shaanxi Province under Grant 2023-JC-ZD-39 and 2024JC-YBMS-520.

Yang You's research group is supported by the NUS Startup Grant (Presidential Young Professorship), the Singapore MOE Tier-1 Grant, the ByteDance Grant, the NUS ARTIC Grant, the Apple Grant, the Alibaba Grant, and the Adobe Gift.

\newpage
\bibliography{custom}

\newpage
\appendix
\onecolumn
\clearpage
\appendix
\section{Dataset Task Distribution}
\label{app:dataset_dist}

GroupToM-Bench comprises 240 unique multi-agent scenarios, evenly distributed across the eight socio-psychological domains (30 scenarios per domain). To probe the complete reasoning arc without introducing class imbalance, each scenario is paired with exactly 13 reasoning tasks: two questions for each of the levels from L1 to L6, and one overarching open-ended question for L7 (Mechanistic Attribution). 

This systematic generation yields a total of 3,120 evaluation tasks. Specifically, each of the eight domains contains exactly 60 tasks for each level from L1 to L6, and 30 tasks for L7. This structure ensures uniform evaluation density across both the horizontal axis (social domains) and the vertical axis (cognitive complexity).

\section{Human Expert Demographics}
\label{app:human_experts}

To ensure the psychological validity and internal consistency of GroupToM-Bench, we recruited a panel of twelve professionals across two strictly separated cohorts, preventing any data contamination between construction and evaluation.

The Generation and Authoring Cohort comprised seven experts responsible for designing seed scenarios, specifying hidden BDI states, and drafting all reasoning tasks and gold-standard references. This group included three Ph.D. holders in Cognitive Science or Social Psychology with specializations in group dynamics, two senior organizational behaviorists, and two AI alignment researchers.

The Verification and Human Baseline Cohort comprised five independent researchers, all holding postgraduate degrees in Psychology, Sociology, or Human-Computer Interaction. This cohort had no access to any generation materials or gold references. They served two roles: answering the final curated questions cold to establish the human performance ceiling reported in Table~1, and conducting blind spot-checks for the LLM-as-a-judge validation.

The strict separation between cohorts was enforced throughout the project. No member of the Generation Cohort participated in baseline evaluation, and no member of the Verification Cohort reviewed any scenario before it was finalized.

\section{Full Case Examples}
\label{app:full_cases}

This section provides three representative scenarios from GroupToM-Bench, covering Belief Evolution and Polarization, Coordination and Dilemma, and Information Distortion. These examples demonstrate the evaluation of the full reasoning arc, from individual mental-state inference to group-level emergence across the seven-level cognitive audit framework.

\clearpage
\includepdf[pages=-, scale=0.8, pagecommand={}]{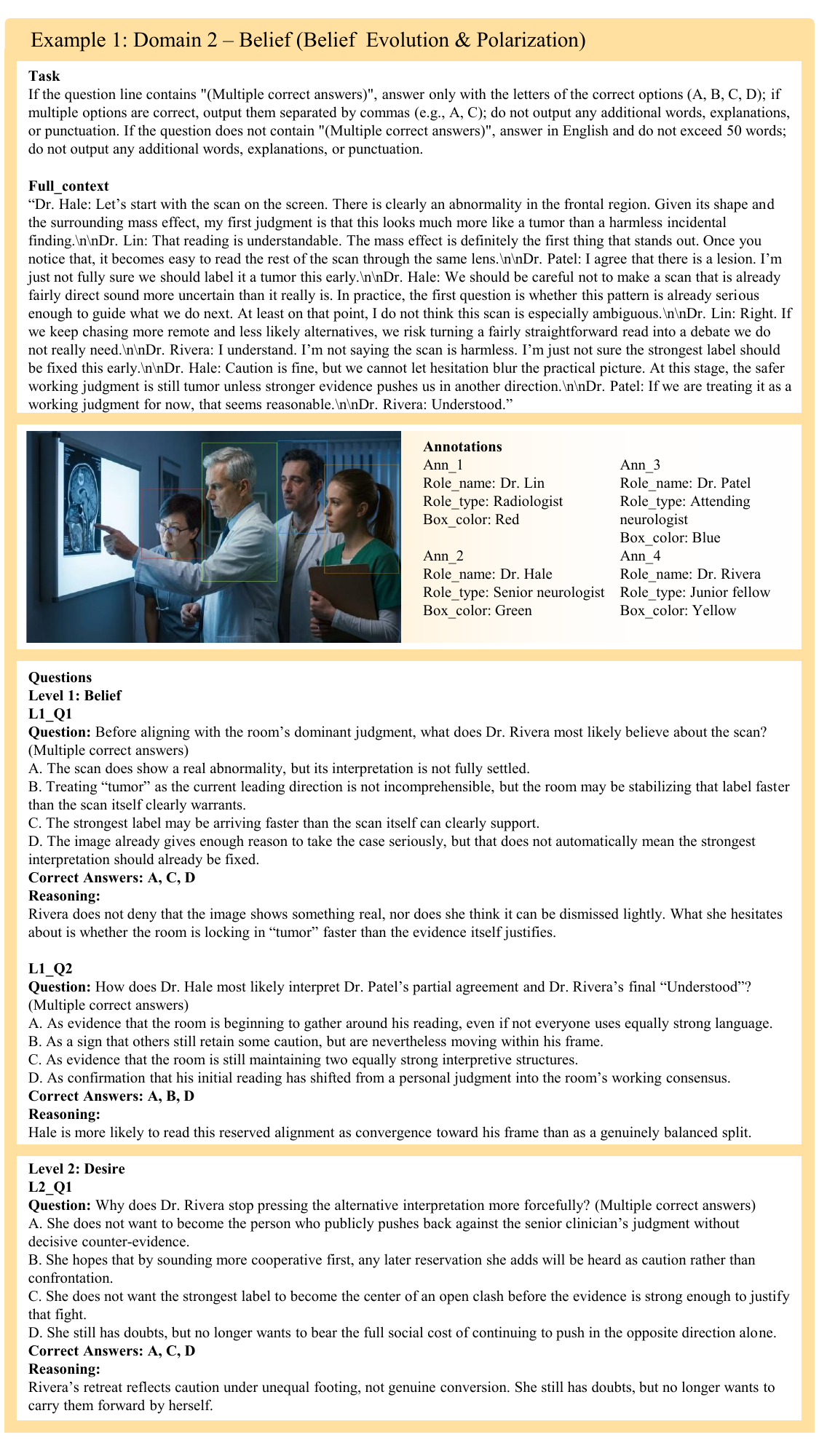}

\clearpage
\includepdf[pages=-, scale=0.8, pagecommand={}]{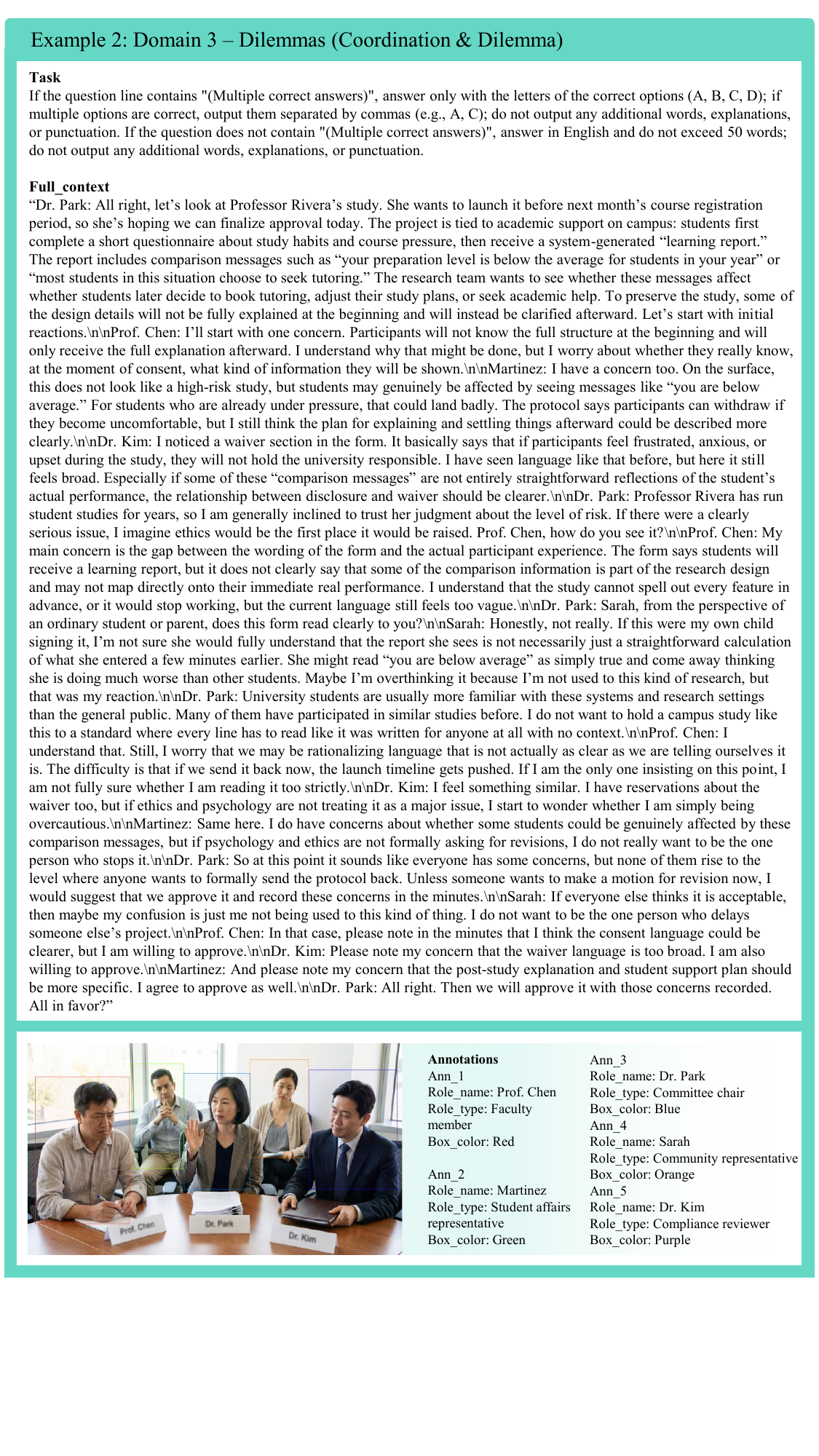}

\clearpage
\includepdf[pages=-, scale=0.8, pagecommand={}]{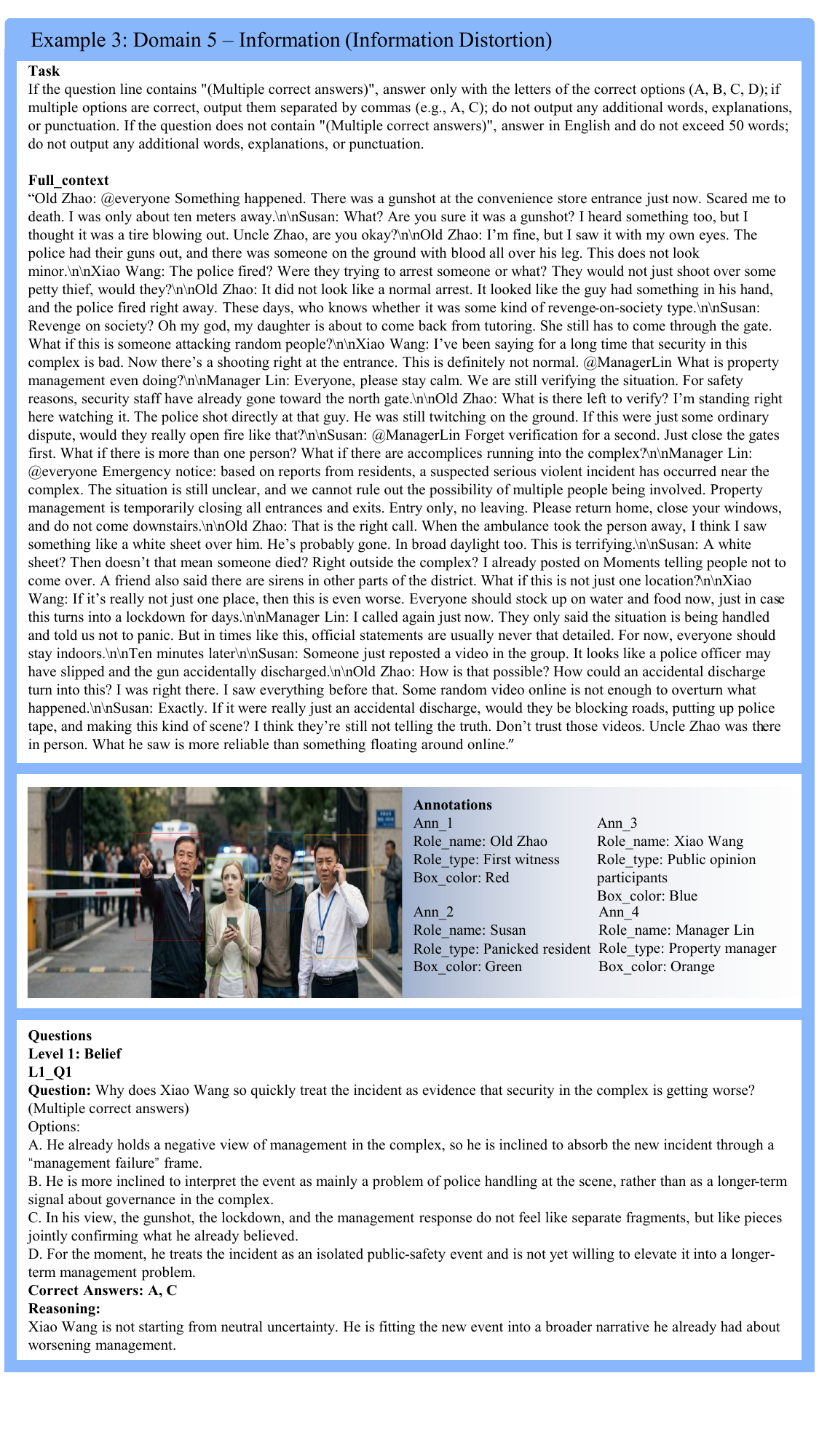}

\end{document}